%% file: main.tex
\documentclass[10pt,twocolumn,letterpaper]{article}

\usepackage{cvpr}              

\input{preamble}

\definecolor{cvprblue}{rgb}{0.21,0.49,0.74}
\usepackage[pagebackref,breaklinks,colorlinks,citecolor=cvprblue]{hyperref}

\def\paperID{216} 
\def\confName{3DV\xspace}
\def\confYear{2024\xspace}

\title{Mixing-Denoising Generalizable Occupancy Networks}

\author{%
  Amine Ouasfi \quad \quad Adnane Boukhayma\\
  Inria, Univ. Rennes, CNRS, IRISA, M2S, France
}

\usepackage{stmaryrd}
\usepackage{amsmath} 
\usepackage{xspace}

\usepackage{algorithm}
\usepackage{algpseudocode}

\makeatletter
\DeclareRobustCommand\onedot{\futurelet\@let@token\@onedot}
\def\@onedot{\ifx\@let@token.\else.\null\fi\xspace}

\def\eg{\emph{e.g}\onedot} 
\def\ie{\emph{i.e}\onedot} 
 
 \def\vs{\emph{vs}\onedot}
\def\wrt{w.r.t\onedot} 
\def\etal{\emph{et al}\onedot}
\makeatother

\begin{document}



\maketitle
\begin{abstract}

While current state-of-the-art generalizable implicit neural shape models \cite{peng2020convolutional,boulch2022poco} rely on the inductive bias of convolutions, it is still not entirely clear how properties emerging from such biases are compatible with the task of 3D reconstruction from point cloud.  We explore an alternative approach to generalizability in this context. We relax the intrinsic model bias (\ie using MLPs to encode local features as opposed to convolutions) and constrain the hypothesis space instead with an auxiliary regularization related to the reconstruction task, \ie denoising. The resulting model is the first only-MLP locally conditioned implicit shape reconstruction from point cloud network with fast feed forward inference. Point cloud borne features and denoising offsets are predicted from an exclusively MLP-made network in a single forward pass. A decoder predicts occupancy probabilities for queries anywhere in space by pooling nearby features from the point cloud order-invariantly, guided by denoised relative positional encoding. We outperform the state-of-the-art convolutional method \cite{boulch2022poco} while using half the number of model parameters.
\end{abstract}




\maketitle

\section{Introduction}
\label{sec:intro}

One of the most sought after goals in modern day computer vision and machine learning is enabling machines to understand and navigate 3D given limited input. This faculty is manifested in several downstream vision and graphics tasks, such as shape reconstruction from noisy and relatively sparse point clouds. Recovering full shape from point clouds is all the more an important problem in account of the ubiquity of this light, albeit incomplete, 3D representation, whether it is acquired from the nowadays increasingly accessible scanning devices, or as obtained from multi-view vision algorithms such as Structure From Motion or Multi-View Stereo (\eg \cite{schoenberger2016sfm,schoenberger2016mvs}). While classical optimization based approaches such as Poisson Reconstruction \cite{kazhdan2013screened} or moving least squares \cite{guennebaud2007algebraic} can mostly successfully deliver from dense clean point sets and normal estimations, the deep learning based more recent alternatives offer faster and more robust prediction especially for noisy and sparse inputs, without requiring normals.  

\begin{figure}[t!]
  \centering
\includegraphics[width=1.0\linewidth]{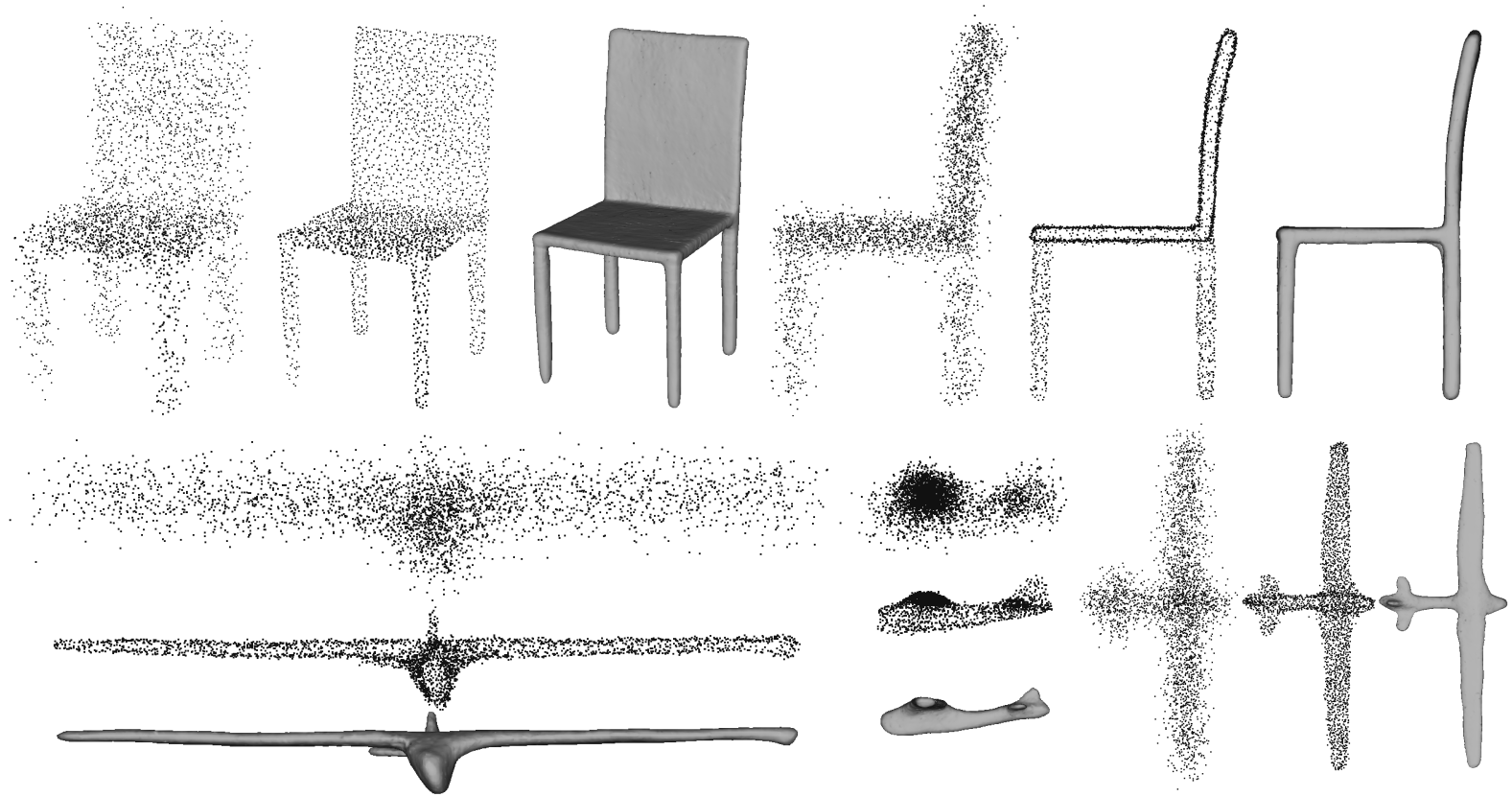}
\caption{Our method couples 3D implicit reconstruction with explicit denoising of point clouds. We show here reconstruction examples from 3000 noisy points (Gaussian noise of std. dev. 0.025). We show the input, our intermediate denoised point cloud, and our predicted geometry.}
\label{fig:den}
\end{figure}

For learning based reconstruction from point cloud, feed forward (optimization-free) generalizable models are important to the community by virtue of their high performances and fast inference. State-of-the-art ones are based on implicit neural shape representations. 
Typically \cite{peng2020convolutional,chibane2020implicit}, a deep convolutional network builds features from the input point cloud, then an implicit decoder maps the feature of a query point to its occupancy or signed distance \wrt the target shape. While the ConvNet predicts explicit extrinsic features in \cite{peng2020convolutional,chibane2020implicit,peng2021shape}, recent work \cite{boulch2022poco} suggests that learning features intrinsically \ie only at the point cloud yields superior results. In this case, features for query points are pooled from the nearest points of the input point cloud. 

While the aforementioned existing work advocates using convolutions to achieve generalizability for this problem, we ask our selves the question, what if we do not constrain our hypothesis space in such an Ad hoc way ? \ie what if we use the most inductive bias free model posssible, and rely on supervision and eventually regularizations, would we succeed in achieving generalization while maintaining fast inference ? 
This line of thought is inspired by the recent success of MLP-only models in vision \cite{tolstikhin2021mlp} and their application to point cloud processing \cite{choe2021pointmixer}. Their idea consists in applying MLPs (mixing) to the data feature dimension and the spatial dimension separately and repeatedly. Point clouds being unordered and irregular, the spatial (Token) mixing is achieved with an order-invariant (Softmax) based pooling on a limited support from the point cloud. 

Based on all the above, we propose a new encoder-decoder model for implicit reconstruction from point cloud. It processes the input point cloud intrinsically (as in \cite{boulch2022poco}) but uses solely MLPs from start to end (unlike \cite{boulch2022poco}). 
This allows us to outperform the state-of-the-art method \cite{boulch2022poco} whilst using only half the number of parameters (6.5M for us, \vs 12.5M for \cite{boulch2022poco}). Not only this shows that convolution inductive biases are not a necessity in this context, but we even achieve a more parameter efficient intrinsic point cloud reconstruction model without them.  Additionally, we show that coupling our implicit reconstruction learning with a related regularization task, in the form of weakly supervised point cloud denoising, can yield even better generalization.  In fact, the performance gap between our method trained with and without denoising increases fourfold in Chamfer distance, when going to a setup requiring more generalization ability (See intra-ShapeNet generalization in Table \ref{tab:sn3k}, \vs the more challenging ShapeNet to ScanNet generalization in Table \ref{tab:scan}).

Our network builds on the dense task prediction model PointMixer \cite{choe2021pointmixer} to predict per-point features. However, to infer occupancies for extrinsic query points (not only on the point cloud), we introduce the new "Extra-set" mixing layer in concordance with PointMixer terminology. The "Extra-set" mixing layer pools features from nearby points of the input point cloud. As this pooling can be guided with positional encoding of the query relative to these support points, we task our PointMixer additionally with denoising the point cloud in parallel, and we use the denoised point cloud to compute more accurate positional encodings (unlike \cite{boulch2022poco}). 
Furthermore, differently from \cite{boulch2022poco}, we incorporate a global decoder using coarse features on a downsampled point cloud as pooling support, in order to robustify the method against local outliers.

We note that existing MLP-only feed forward generalizable methods lack a local feature aggregation mechanism in their encoder architectures (PointNet), which is crucial for performance without sacrificing inference speed. \cite{mescheder2019occupancy} used a global feature causing under-performance. \cite{erler2020points2surf} enforces locality through query point dependent local patch inputs, hence requiring as many encoder forward passes as there are query points. This naturally results in significantly increased inference times surpassing even \eg some auto-decoding optimization based methods. Conversely, being equipped with point mixing locality, our MLP-only model's encoder solely requires a single pass for the input point cloud. It offers consequently about 3 orders of magnitude faster reconstruction (within 500ms at resolution $128^3$) compared to \cite{erler2020points2surf}, in line with convolutional locality endowed competitors (\cite{boulch2022poco,peng2020convolutional}).

We outperform existing comparable methods on standard benchmarks for object and scene level reconstruction, and in generalization to real scan data (Section \ref{sec:res}). The performance gap is most important in the most sparse and noisy input setup (Table \ref{tab:sn300}), demonstrating our resilience to scarcity and corruption. Our reconstructions are also more robust, as witnessed by our superior L2 Chamfer errors across all benchmarks. We evaluate the benefits of each of our contributions in our ablation section (\ref{sec:abl}).

Our contributions can be summarized as follows:\\
\noindent $\bullet$ A convolution-free feed forward intrinsic occupancy network from point cloud attaining the new state-of-art, while being twice as parameter efficient as its closest intrinsic convolutional counterpart (\cite{boulch2022poco}).\\
\noindent $\bullet$ Joint learning of implicit reconstruction from point cloud with explicit point cloud denoising for the first time to the best of our knowledge.\\
\noindent $\bullet$ Combining local and global pooling based decoding for deep intrinsic (\eg \cite{boulch2022poco}) occupancy regression.

\section{Related work}
\label{sec:related}

\paragraph{Shape Representations in Deep Learning}
Shapes can be represented in deep learning either intrinsically or extrinsically. Intrinsic representations discritize the shape itself. When done explicitly using \eg tetrahedral or polygonal meshes \cite{wang2018pixel2mesh,kato2018neural}, or point clouds \cite{fan2017point}, this bounds the output topology thus limiting the variability of outputs. Among other forms of intrinsic representations, 2D patches \cite{groueix2018papier,williams2019deep,deprelle2019learning} can prompt discontinuities, whilst the simple nature of shape primitives such as cuboids \cite{abstractionTulsiani17,zou20173d}, planes \cite{liu2018planenet} and Gaussians \cite{genova2019learning} limits their expressivity. Differently, extrinsic shape representations model the space containing the scene/object. Voxel grids \cite{wu20153d,wu2016learning} are the most popular one being the extension of 2D pixels to 3D domain. Their capacity is limited though by their cubic resolution memory cost. Sparse representation like octrees \cite{riegler2017octnet,tatarchenko2017octree,wang2017cnn} can alleviate this issue to some extent. 

\paragraph{Implicit Neural Shape Representations}
Implicit neural representations (INRs) stood out as a major new medium for modelling shape and radiance (\eg \cite{mildenhall2020nerf,yariv2021volume,wang2021neus,jain2021dreamfields,chan2022efficient}) extrinsically. They overcome many of the limitations of the aforementioned classical representations thanks to their ability to represent shapes with arbitrary topologies at virtually infinite resolution. They are usually parameterised with MLPs mapping spatial locations or features to \eg occupancy \cite{mescheder2019occupancy}, signed \cite{park2019deepsdf} or unsigned \cite{chibane2020neural,Zhou2022CAP-UDF} distances relative to the target shape. The level-set of the inferred field from these MLPs can be rendered through ray marching \cite{hart1996sphere}, or tessellated into an explicit shape using \eg Marching Cubes \cite{lorensen1987marching}. A noteworthy branch of work builds hybrid implicit/explicit representations \cite{palmer2022deepcurrents,chen2020bsp,deng2020cvxnet,yavartanoo20213dias} based mostly on differentiable space partitioning. 

\paragraph{Generalizing Implicit Neural Shape Representations}
In order to represent collections of shape, implicit neural models require conditioning mechanisms. These include latent code concatenation, batch normalization, hypernetworks \cite{sitzmann2020implicit,NEURIPS2019_b5dc4e5d,sitzmann2021light,wang2021metaavatar} and gradient-based meta-learning \cite{ouasfi2022few,sitzmann2020metasdf}. Concatenation based conditioning was first implemented using single global latent codes \cite{mescheder2019occupancy,chen2019learning,park2019deepsdf}, and further improved with the use of local features \cite{li2022learning,genova2020local,tretschk2020patchnets,takikawa2021neural,peng2020convolutional,chibane2020implicit,jiang2020local,erler2020points2surf}. 

\paragraph{Shape Reconstruction from Point Cloud}
Classical approaches include combinatorical ones where the shape is defined through an input point cloud based space partitioning, through \eg alpha shapes \cite{bernardini1999ball} Voronoi diagrams \cite{amenta2001power} or triangulation \cite{cazals2006delaunay,liu2020meshing,rakotosaona2021differentiable}. On the other hand, the input samples can be used to define an implicit function whose zero level set represents the target shape, using global smoothing priors \cite{williams2022neural,lin2022surface,williams2021neural} \eg radial basis function \cite{carr2001reconstruction} and Gaussian kernel fitting \cite{scholkopf2004kernel}, local smoothing priors such as moving least squares \cite{mercier2022moving,guennebaud2007algebraic,kolluri2008provably,liu2021deep}, or by solving a boundary conditioned Poisson equation \cite{kazhdan2013screened}. The recent literature proposes to parameterise these implicit functions with deep neural networks and learn their parameters with gradient descent, either in a supervised or unsupervised manner.

\paragraph{Unsupervised Implicit Neural Reconstruction}
A neural network is typically fitted to the input point cloud without extra information. Regularizations can improve the convergence such as the spatial gradient constraint based on the Eikonal equation introduced by Gropp \etal \cite{gropp2020implicit}, a spatial divergence constraint as in \cite{ben2022digs}, Lipschitz regularization on the network \cite{liu2022learning}. Atzmon \etal learn an SDF from unsigned distances \cite{atzmon2020sal}, and further supervises the spatial gradient of the function with normals \cite{atzmon2020sald}. Ma \etal \cite{ma2020neural} expresses the nearest point on the surface as a function of the neural signed distance and its gradient. They also leverage 
self-supervised local priors to deal with very sparse inputs \cite{ma2022reconstructing} and improve generalization \cite{ma2022surface}. All of the aforementioned work benefits from efficient gradient computation through back-propagation in the neural network. Periodic activations were introduced in \cite{sitzmann2020implicit}. Lipman \cite{lipman2021phase} learns a function that converges to occupancy while its log transform converges to a distance function. \cite{williams2021neural} learns infinitely wide shallow MLPs as random feature kernels.  

\begin{figure*}[th!]
  \centering
\includegraphics[width=1.0\linewidth]{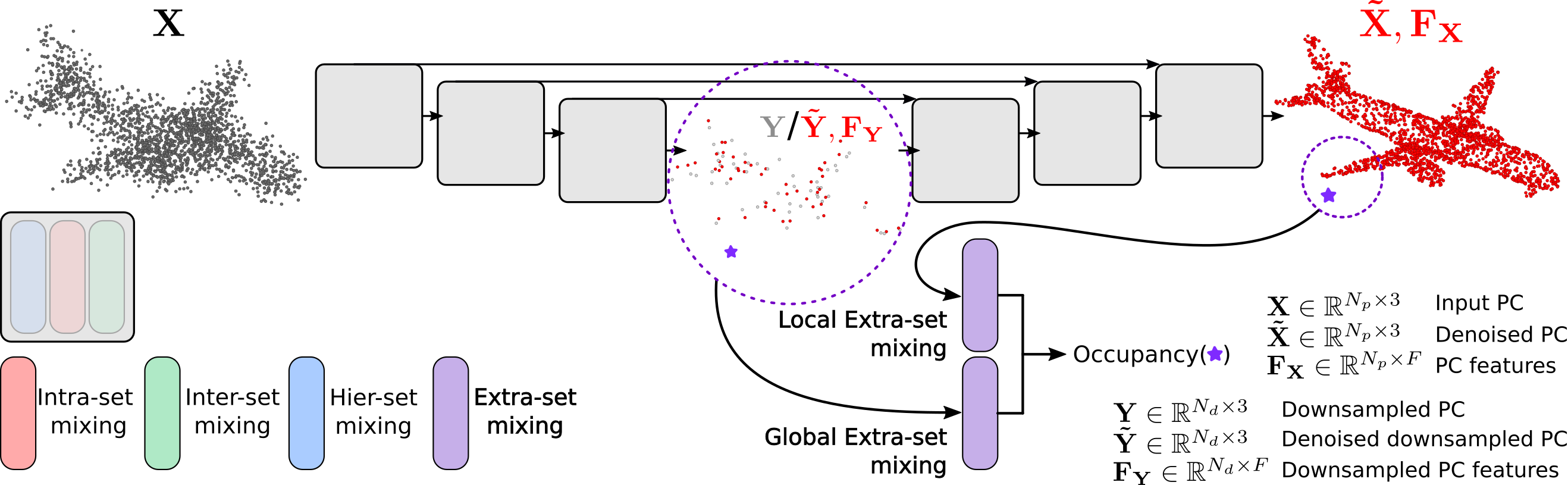}
\caption{Overview: Our method uses exclusively MLPs from beginning to end to predict an implicit shape function given a noisy input point cloud. First a U-Net PointMixer \cite{choe2021pointmixer} (Gray boxes) is tasked with denoising the input and producing per point features. The building blocks of this network combine 3 pre-exiting forms of point mixing which pool features from support points in a point cloud into other points of the point cloud. Channel-wise mixing is also used throughout the model as well. We refer the reader to the work of Choe \etal \cite{choe2021pointmixer} for a detailed description on these preexisting mixing layers. We introduce "Extra-set mixing" (Section \ref{sec:mix}), which we use to pool features both locally and globally onto any external query point to estimate its occupancy. This mixing is guided by relative positional encoding using the denoised support point coordinates (See Figure \ref{fig:mix}).}
\label{fig:pipe}
\end{figure*}

\paragraph{Supervised Implicit Neural Reconstruction}
Supervised methods assume a labeled training data corpus commonly in the form of dense samples with ground truth shape information. Auto-decoding methods \cite{li2022learning,park2019deepsdf,tretschk2020patchnets,jiang2020local,chabra2020deep} require test time optimization to be fitted to a new point cloud, which can take up to several seconds. Encoder-decoder based methods enable fast feed forward inference. Introduced first in this respect, Pooling-based set encoders  \cite{mescheder2019occupancy,chen2019learning,genova2020local} such as PointNet \cite{qi2017pointnet} have been shown to underfit the context. Convolutional encoders yield state-of-the-art performances. They use local features either defined in explicit volumes and planes \cite{peng2020convolutional,chibane2020implicit,lionar2021dynamic,peng2021shape} or solely at the input points \cite{boulch2022poco}. Ouasfi and Boukhayma \cite{ouasfi2023Robustifying} proposed concurrently to robustify the generalization of these models through transfer learning to a kernel ridge regression whose hyperparameters are fitted to the shape. Peng \etal \cite{peng2021shape} proposed a differentiable Poisson solving layer that converts predicted normals into an indicator function grid efficiently. However, it is limited to small scenes due to the cubic memory requirement in grid resolution.   


\section{Method}
\label{sec:method}

Given a noisy input point cloud $\mathbf{X}\subset{\mathbb R}^{ N_p\times 3}$, our objective is to recover a shape surface $\mathcal{S}$ that best explains this observation, \ie the input point cloud elements being noisy samples from $\mathcal{S}$. 

To achieve this, we train a deep implicit neural network $f_{\theta}$ to predict occupancy values relative to a target shape $\mathcal{S}$ at any queried Euclidean space location $x \in {\mathbb R}^{3}$, given the input point cloud $\mathbf{X}$, \ie $f_{\theta}(\mathbf{X},x)=1$ if $x$ is inside, and $0$ otherwise. The inferred shape $\mathcal{\hat{S}}$ can then be obtained as a level set the occupancy field inferred using $f_{\theta}$:
\begin{equation}
\hat{\mathcal{S}} = \{ x\in\mathbb{R}^3 \mid f_{\theta}(\mathbf{X},x) = 0.5\}.
\end{equation}
In practice, an explicit triangle mesh for $\mathcal{\hat{S}}$ is extracted using the Marching Cubes \cite{lorensen1987marching} algorithm.  

We opt for a feed forward conditional implicit model $f_{\theta}$, as we want our method to generalize to multiple shapes simultaneously and provide fast test-time optimization free inference. Similarly to the state-of-the-art in this category \eg \cite{boulch2022poco,peng2020convolutional}, $f_{\theta}$ consists of a point cloud conditioning network and an implicit decoder.   
  
\subsection{Model}

The staple back-bone model for feed-forward generalizable 3D shape reconstruction from point cloud has lately been the one introduced by Peng \etal \cite{peng2020convolutional}, which also bears similarities with the concurrent work by Chibane \etal \cite{chibane2020implicit}. It has been widely used among methods predicting other implicit fields from point clouds as well, such as point displacements \cite{venkatesh2021deep}, unsigned distances \cite{chibane2020neural} or whether point pairs are separated by a surface \cite{ye2022gifs}. In this model, the encoder builds an explicit extrinsic 3D feature volume from the voxelized or pixelized input point cloud through 3D (or 2D) convolutional networks.     

Recently, Boulch \etal \cite{boulch2022poco} argued against this  strategy. They contented that these methods operate in a voxelized discretization whose voxel centers maybe far from the input point cloud locations. They also argued that
voxel centers holding the shape information are uniformly sampled, while they should be ideally more focused near the surface. Conclusively, they introduced a point cloud convolutional encoder, where features are only defined at the input point cloud locations, where shape information matters most. They achieve superior reconstruction performances consequently on several benchmarks.   

Both of the aforementioned strategies however still rely on convolutions for translational equivariance and generalization across scenes and objects.  Differently, we propose here an architecture based solely on MLPs, and show that it mostly outperforms the previously mentioned strategies. 

\subsection{Feature Extraction and Denoising Network}

Our point cloud feature extraction network takes noisy point cloud $\mathbf{X}$ as input and produces an output per element in the point cloud. However, differently from \cite{boulch2022poco}, it does not use convolutions, but bases the entire model on MLPs instead, inspired by recent success of the PointMixer \cite{choe2021pointmixer} architecture.

Specifically, we use the dense prediction task model in \cite{choe2021pointmixer}, which consists of a U-Net \cite{ronneberger2015u} symmetric encoder-decoder scheme. The point cloud is downsampled gradually throughout the encoder $E$ using Farthest point sampling until it reaches the coarsest point cloud $\mathbf{Y} \in \mathbb{R}^{N_d\times 3}$. It is upsampled back then correspondingly through out the decoder $D$. As illustrated in Figure \ref{fig:pipe}, each block in the encoder/decoder consists mostly of a Hierarchical mixing layer, an intra-set mixing layer, a inter-set mixing layer and channel mixing, as elaborated in \cite{choe2021pointmixer}.

The channel mixing layer is an MLP operating directly on the feature domain, \ie mixing feature channels. The other mixings are point-wise mixings to replace the token-wise mixing in the original MLP-Mixer model \cite{tolstikhin2021mlp}. They are softmax based order invariant pooling layers, that pool features from a limited set of point cloud support points into a point cloud query. They vary in the way their supports are defined. In the intra-set one the support is the nearest points to the query. In the inter-set one the support is the inverse of the intra-set support, \ie a point is part of the support of a query if the query is part of the nearest points of that point. The hierarchical-set mixing is used for down/up transitions in the encoder/decoder, and hence the support is the nearest points in the previous/next point cloud resolution. In practice, for the decoder to be symmetrical with the encoder, the up transition support set is defined as the inverse of the equivalent down transition support set, differently from seminal work \cite{zhao2021point,zhao2019pointweb,qi2017pointnet++}. For a more thorough explanation of the intra-set, inter-set and hierarchical-set mixing layers we refer the reader to the work by Choe \etal \cite{choe2021pointmixer}.   

Differently from \cite{boulch2022poco}, we design our feature extraction network to perform two tasks simultaneously: produce per point fine features
$\mathbf{F}_\mathbf{X} \in \mathbb{R}^{N_p \times F}$ and displacement vectors $\mathbf{\Delta X} \in \mathbb{R}^{N_p\times 3}$:
\begin{equation}
    \mathbf{F_X},\mathbf{\Delta X} = D\circ E(\mathbf{X}). 
\end{equation}

We train the displacements semi-supervisedly to produce a denoised version $\mathbf{\tilde{X}} \in \mathbb{R}^{N_p \times 3}$ of the input point cloud:
\begin{equation}
    \mathbf{\tilde{X}} = \mathbf{X} + \mathbf{\Delta X}.  
\end{equation}

We note that using the encoder $E$, we can also extract coarse features $\mathbf{F_Y}$ for the downsampled point cloud $\mathbf{Y}$ at the bottleneck of the model:
\begin{equation}
    \mathbf{F_Y} = E(\mathbf{X}).   
\end{equation}

A denoised version $\mathbf{\tilde{Y}} \in \mathbb{R}^{N_d \times 3}$ of the downsampled input point cloud $\mathbf{Y}$ can be naturally obtained from the output denoised point cloud $\mathbf{\tilde{X}}$ via the encoder defined downsampling indexing $\delta: \llbracket 1,N_p\rrbracket \rightarrow \llbracket 1,N_d\rrbracket$:

\begin{equation}
    \mathbf{\tilde{Y}}_i = \mathbf{\tilde{X}}_{\delta(i)},\quad  \mathbf{Y}_i = \mathbf{X}_{\delta(i)}.    
\end{equation}

\subsection{Extra-set Mixing Decoder}
\label{sec:mix}

To serve the role of our implicit decoder, we introduce a new point mixing layer, borrowing the terminology in PointMixer \cite{choe2021pointmixer}. As opposed to the existing point mixing layers in \cite{choe2021pointmixer}, this layer takes as input a query point representing any location in Euclidean space, and not necessarily an element from the input point cloud topology $\mathbf{X}$, hence the denomination "Extra-set". Similarly to the point-mixing layers in \cite{choe2021pointmixer}, it pools features from a support set using softmax in an order-invariant fashion.

\begin{figure}[t!]
\centering
\includegraphics[width=\linewidth]{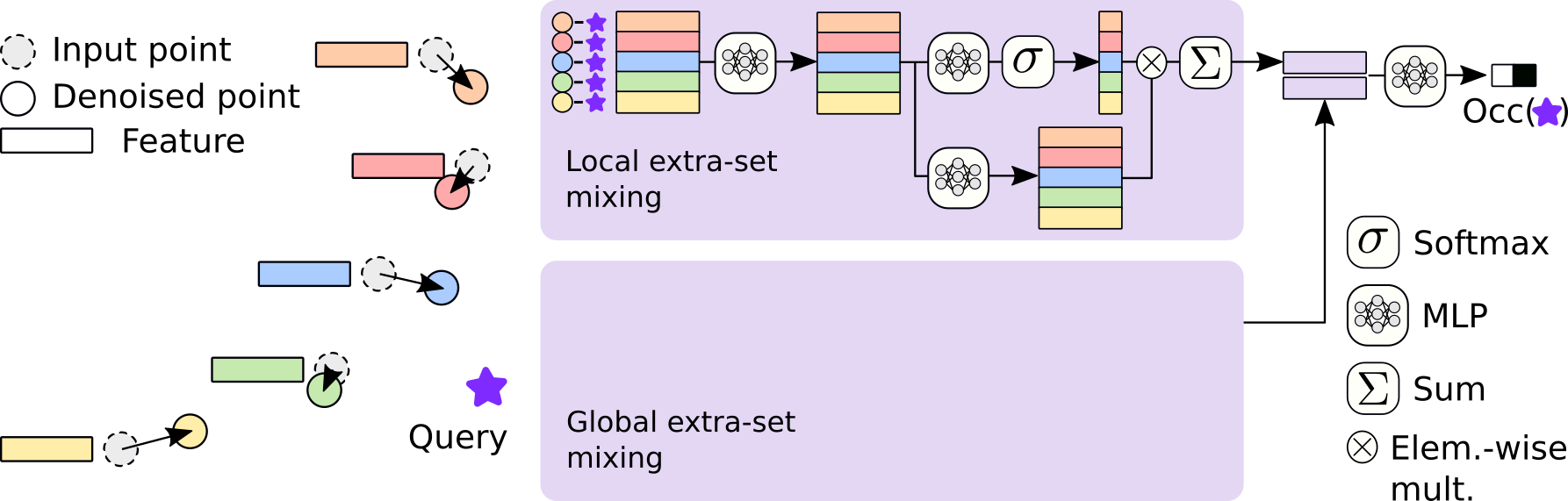}
\caption{Using denoised point coordinates and features, our implicit decoder performs local and global extra-set point mixing at a given query location to predict its shape occupancy value. The global extra-set mixing mechanism is similar to the local one. The local extra-set mixing uses the $K$ nearest denoised points of the point cloud. The global extra-set mixing uses to the full $N_d$ denoised points of the downsampled point cloud.}
\label{fig:mix}
\end{figure}

We first use our layer to extract a local fine feature $\mathbf{y}^{loc}$ for a query point $q\in \mathbb{R}^3$. 
This feature is built using the $K$ nearest points to $q$ from the denoised output point cloud $\mathbf{\hat{X}}$ and their corresponding output fine features in $\mathbf{F_X}$, as generated by decoder $D$.

To this end, and as illustrated in Figure \ref{fig:mix}, we first compute an aggregation score $s_{\tilde{p}}$ for each support element $\tilde{p}$:
\begin{equation}
    s_{\tilde{p}} = g_2 \circ g_1 (\tilde{p}-q;{\mathbf{F}_{\mathbf{X}}}(p)),\quad \tilde{p}\in \mathcal{X}_q= k\text{NN}(\mathbf{\tilde{X}},q),
\end{equation}
where $g_1(.)$ and $g_2(.)$ are channel mixing MLPs and $\mathbf{F_X}(p)$ is the fine feature of denoised support point $\tilde{p}$. We use these aggregation scores to pool features of the support points into the final local feature of the query point:
\begin{equation}
    \mathbf{y}^{\text{loc}} = \sum_{\tilde{p}\in \mathcal{X}_q} \text{softmax}(s_{\tilde{p}}) \odot \bigl[
    g_3 \circ g_1(\tilde{p}-q;{\mathbf{F}_{\mathbf{X}}}(p))
    \bigr],
\end{equation}
where $\text{softmax}(.)$ is the softmax normalization over the support point dimension, $g_3(.)$ is another channel mixing MLP, and $\odot$ denotes the element wise product. Conclusively, this layer is in fact akin to a spatial attention module \cite{woo2018cbam}, space here spanning the support points.  

We note additionally that the architecture of this layer follows the decoder in \cite{boulch2022poco}. Most importantly, we pinpoint that a main difference here is that the relative positional encoding in our extra-set mixing uses denoised support coordinates: $\{\tilde{p}-q\}$, $\tilde{p}\in \mathbf{\tilde{X}}$, instead of the original noisy ones: $\{p-q\}$, $p\in \mathbf{X}$, as done in \cite{boulch2022poco}. This allows us to reduce the effect of noise on the Euclidean relative position guided feature aggregation. 
Similarly to \cite{boulch2022poco}, we use in practice $N_H$ MLPs $g_2$ in parallel, \ie  $\text{\small{softmax}}(s_{\tilde{p}}):= \frac{1}{N_H}\sum_{h=1}^{N_H} \text{\small{softmax}}(g_2^h \circ g_1(\tilde{p}-q;\mathbf{F}(p)))$.

Next, we use our layer introduced above to extract a global coarse feature $\mathbf{y}^{glob}$ as well for a query point $q\in \mathbb{R}^3$. The goal of this global mixing component is to include global context in the reconstruction task and prevent the method from over-fitting on local information. This feature is built using the entire $N_d$ sized down-sampled denoised point cloud $\mathbf{\tilde{Y}}$ elements, along with their corresponding output coarse features in $\mathbf{F_Y}$, as generated by encoder $E$.     

Similarly to the previous case, we first compute aggregation scores $r_{\tilde{p}}$ for each support element $\tilde{p}$, and we use a support point-wise softmax normalization of these scores to weight the support point features: 
\begin{gather}
    r_{\tilde{p}} = h_2 \circ h_1 (\tilde{p}-q;{\mathbf{F}_{\mathbf{Y}}}(\tilde{p})),\quad \tilde{p}\in \mathbf{\tilde{Y}}.\\
    \mathbf{y}^{\text{glob}} = \sum_{\tilde{p}\in \mathbf{\tilde{Y}}} \text{softmax}(r_{\tilde{p}}) \odot \bigl[h_3 \circ h_1(\tilde{p}-q;{\mathbf{F}_{\mathbf{Y}}}(\tilde{p}))\bigr],
\end{gather}
where $h_1(.)$, $h_2(.)$ and $h_3(.)$ are channel mixing MLPs, and $\mathbf{F_Y}(p)$ is the coarse feature of denoised support point $\tilde{p}$. 

Finally, a last channel mixing MLP $g_4(.)$ combines the local feature $\mathbf{y}^{loc}$ and the global feature $\mathbf{y}^{glob}$ of a query point $q$ to produce the occupancy probability of the latter, which writes:
\begin{equation}
    f_{\theta}(\mathbf{X},q) := \text{occ}(q) = g_4(\mathbf{y}^{\text{loc}};\mathbf{y}^{\text{glob}}).
\end{equation}

\subsection{Training}

Our method is fully differentiable and is trained end-to-end leveraging the combination of a reconstruction loss and a denoising loss: 
\begin{equation}
    \mathcal{L} = \mathcal{L}_{\text{rec}} + \mathcal{L}_{\text{den}}.
\end{equation}

\begin{figure}[t!]
  \centering
\includegraphics[width=1.0\linewidth]{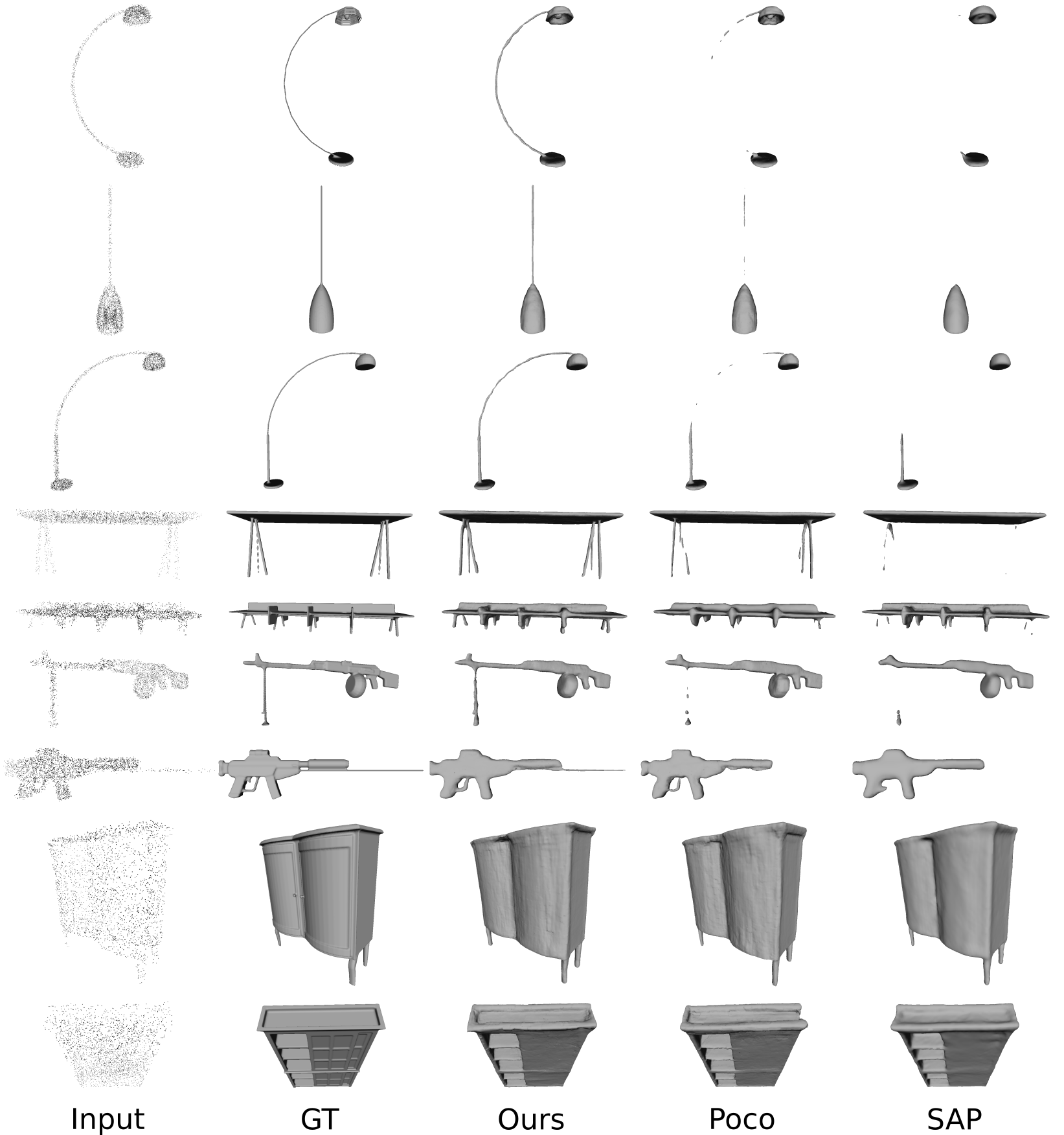}
\caption{ShapeNet reconstructions from 3000 noisy points, with standard deviation of 0.005. Our method reproduces details and fine structures with more fidelity.}
\label{fig:sn3k5}
\end{figure}

We train the feature extraction and denoising network to produce a denoising offset for each input point by back-propagating the L2 Chamfer distance between the denoised point cloud $\mathbf{\tilde{X}}$ and a ground truth point cloud sampled from the ground truth shape $\mathbf{X}_{\text{gt}}\sim \mathcal{S}$:
\begin{equation}
    \mathcal{L}_{\text{den}} =
    \frac{1}{2|\mathbf{\tilde{X}}|}\sum_{\tilde{p}\in\mathbf{\tilde{X}}}\min_{p\in\mathbf{X}_{\text{gt}}}\Vert p-\tilde{p} \Vert_2^2
    + \frac{1}{2|\mathbf{X}_{\text{gt}}|}\sum_{p\in\mathbf{X}_{\text{gt}}}\min_{\tilde{p}\in\mathbf{\tilde{X}}}\lVert \tilde{p}-p \rVert_2^2,
\end{equation} 
where $|.|$ is the point set cardinality. A similar loss is used in recent supervised point cloud denoising literature (\eg \cite{rakotosaona2020pointcleannet}). Ground truth point cloud $\mathbf{X}_{\text{gt}}$ counts 100k ground truth surface samples. Besides providing more accurate support relative positional encoding for the decoder extra-set mixing, this loss can provide additional regularization to the feature extraction network, due to the correlations between the denoising and reconstruction tasks. 

Following seminal work \cite{mescheder2019occupancy}, the reconstruction loss is the Binary Cross-entropy loss between query points $\{q\}$ sampled around the ground truth surface $\mathcal{S}$ and their ground truth occupancy labels: 
\begin{equation}
\mathcal{L}_{\text{rec}} = \sum_q \text{BCE}(\text{occ}(q),\text{occ}(q)_{\text{gt}}).
\end{equation}

\section{Results}
\label{sec:res}

We compare in this section our approach quantitatively and qualitatively to state-of-the-art methods on object and scene reconstruction benchmarks, using both synthetic and real data. 

\subsection{Implementation details}
We implement our method using the PyTorch Framework. The U-Net follows the architecture in \cite{choe2021pointmixer}. Coarse and fine feature dimensions are $F_c$ = 512 and  $F_f$ = 32 respectively.  
For inputs of size $N_p$ = 3000, we use $N$ = 64 nearest neighbor as the local decoder support size. For $N$ = 300, $N$ = 12. The downsampled point cloud is of size $N_d$ = 12. MLPs $g_1$/$h_1$ have 3 ReLU activated hidden layers of dimension 32. $g_2$/$h_2$ and $g_3/h_3$ have one layer of dimension 64 and 32 respectively. $g_4$ has two layers of dimensions 64 and 2. $N_H$ = 64. Following \cite{boulch2022poco}, we train for 600k iterations in batches of $16$ shapes and $2048$ query points, and we reconstruct similarly at resolution 256. 

\begin{figure}[t!]
  \centering
\includegraphics[width=1.0\linewidth]{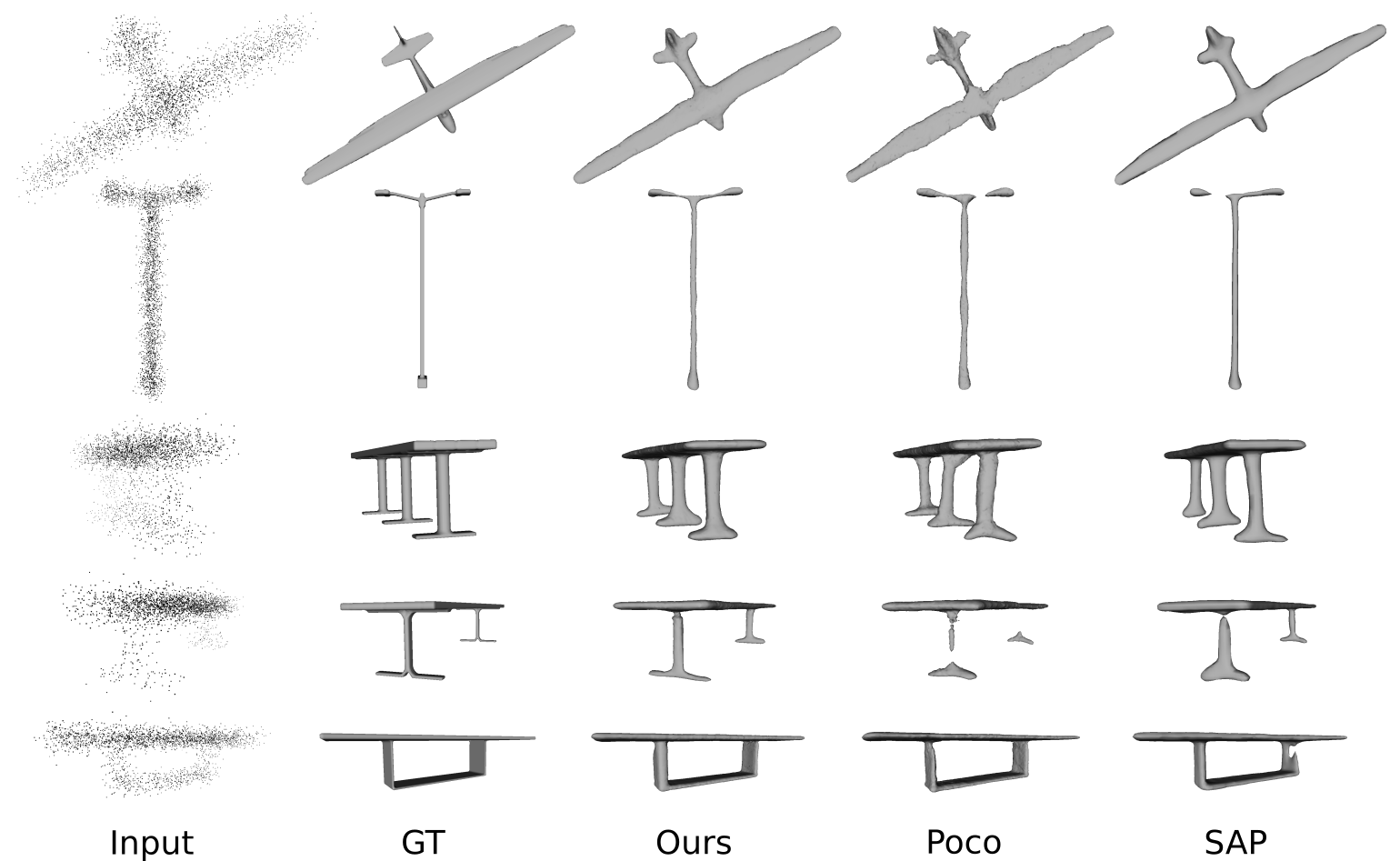}
\caption{ShapeNet reconstructions from 3000 noisy points, with standard deviation of 0.025. We show more resilience to high levels of noise.}
\label{fig:sn3k25}
\end{figure}

\subsection{Metrics} 
Following seminal work, we evaluate our method and the competition \wrt the ground truth using standard metrics for the 3D reconstruction task. Namely, the L1 and L2 Chamfer Distances $\text{CD}_\text{1}\times10^{2}$ and $\text{CD}_\text{2}\times10^4$, Normal Consistency (NC) and the F-Score (FS) based on Euclidean distance. We detail their expressions in the supplementary material.  

\subsection{Datasets}
\textbf{ShapeNet} \cite{shapenet} is used to evaluate object level reconstruction. It consists of various instances of 13 different object classes. We train/test on this dataset using input point clouds of sizes 300 and 3000, and noises of standard deviation ($\sigma$) 0.005 and 0.025. \textbf{Synthetic Rooms} \cite{peng2020convolutional}
is used to evaluate scene level reconstruction. Numerical and qualitative results for this dataset are shown in the supplementary material. It consists of 5000 scenes made of a floor and walls and populated with ShapeNet objects. 
We train/test on this dataset using inputs of size 10k and a noise of standard deviation 0.005. For both datasets, we follow the train/test splits, the training procedure and the evaluation protocol in \cite{boulch2022poco}. Finally, we test the ShapeNet trained models on real scan datasets to evaluate generalization. We use the \textbf{FAUST} \cite{Bogo:CVPR:2014} dataset. It consists of real scans of 10 human body identities in 10 different poses. We sample 3000 points from the scans as inputs. We use the provided mesh registrations to compute IoU.
We also use the \textbf{ScanNet v2} \cite{dai2017scannet} dataset. It contains 1513 rooms captured with an
RGB-D camera. We sample 10k points as inputs.

\subsection{Object level reconstruction}

\begin{table}[t!]
\centering
\scalebox{0.7}{
\begin{tabular}{lcccc|cccc}
\toprule
\multicolumn{1}{c}{} & \multicolumn{4}{c}{$\sigma$ = 0.005} & \multicolumn{4}{c}{$\sigma$ = 0.025}\\
\midrule
\multicolumn{1}{c}{} & $\text{CD}_\text{1}$ $\downarrow$ & $\text{CD}_\text{2}$ $\downarrow$ & NC $\uparrow$ & \multicolumn{1}{c}{FS $\uparrow$}  & $\text{CD}_\text{1}$ $\downarrow$ & $\text{CD}_\text{2}$ $\downarrow$ & NC $\uparrow$ & FS $\uparrow$\\
\midrule
SAP \cite{peng2021shape}             & 0.58 & 1.04 &  0.89 & 0.87   & 1.02   & 3.50 &\textbf{0.85} &\textbf{0.69} \\
POCO \cite{boulch2022poco}              & 0.59 & 1.54 &  0.88 & 0.88 & 1.10   & 4.40  & 0.83 & 0.66 \\
\midrule
Ours w/o den.   & 0.55 & \textbf{0.92} &   0.89 & \textbf{0.89}  & 1.05   & 3.56   & 0.84 & 0.66 \\ 
Ours              & \textbf{0.54} & 0.94 & \textbf{0.90} & \textbf{0.89}  & \textbf{1.00} & \textbf{3.49} & \textbf{0.85} & \textbf{0.69}\\
\bottomrule
\end{tabular}}
\caption{ShapeNet reconstruction from 300 noisy points, with standard deviations ($\sigma$) of 0.005 and 0.025.}
\label{tab:sn300}
\end{table}

\begin{table}[t!]
\centering
\scalebox{0.7}{
\begin{tabular}{lcccc|cccc}
\toprule
\multicolumn{1}{c}{} & \multicolumn{4}{c}{$\sigma$ = 0.005} & \multicolumn{4}{c}{$\sigma$ = 0.025}\\
\midrule
\multicolumn{1}{c}{} & $\text{CD}_\text{1}$ $\downarrow$ & $\text{CD}_\text{2}$ $\downarrow$ & NC $\uparrow$ & \multicolumn{1}{c}{FS $\uparrow$} & $\text{CD}_\text{1}$ $\downarrow$ & $\text{CD}_\text{2}$ $\downarrow$ & NC $\uparrow$ & FS $\uparrow$\\
\midrule
SPR \cite{kazhdan2013screened} & 2.98 & - & 0.77 & 0.61 & 4.99 & - & 0.60 & 0.32 \\
3D-R2N2 \cite{choy20163d} & 1.72 & - & 0.71 & 0.40 & 1.72 & - & 0.71 & 0.42 \\
AtlasNet \cite{groueix2018papier} & 0.93 & - & 0.85 & 0.71 & 1.17 & - & 0.82 & 0.53 \\
ConvONet \cite{peng2020convolutional}            & 0.44 & - & 0.94 & 0.94 & 0.66 & - & \textbf{0.91} & 0.85\\
SAP \cite{peng2021shape}                  & 0.34 & 0.60 &  0.94 & 0.97  & \textbf{0.55} & 0.95   & \textbf{0.91} & \textbf{0.89} \\
POCO  \cite{boulch2022poco}                & 0.30 & 0.31 & \textbf{0.95} & 0.98 & 0.58   & 1.24  & 0.90 & 0.88 \\
\midrule
Ours w/o den.   & 0.32 & 0.25 &  0.94 & 0.98 & 0.58   & 0.98   & 0.90 & 0.87 \\ 
Ours                   & \textbf{0.29} & \textbf{0.15} &  \textbf{0.95} & \textbf{0.99} & \textbf{0.55}   & \textbf{0.93}  & \textbf{0.91} & \textbf{0.89}\\
\bottomrule
\end{tabular}}
\caption{ShapeNet reconstruction from 3000 noisy points, with standard deviations ($\sigma$) of 0.005 and 0.025.}
\label{tab:sn3k}
\end{table}

Table \ref{tab:sn300} shows numerical evaluations of ShapeNet \cite{shapenet} reconstructions from 300 noisy input points, with both noises of 0.005 and 0.025 standard deviation. Table \ref{tab:sn3k} shows results for 3000 points with the same noises.
We show numbers for our method (Ours) and our method without denoising (Ours w/o denois.). We report numbers 
for \cite{peng2021shape} (SAP) and \cite{boulch2022poco} (POCO) using their available models when applicable, and train the rest when needed following their official implementations. We compile numbers for \cite{peng2020convolutional} (ConvONet), 
\cite{kazhdan2013screened} (SPR), \cite{choy20163d} (3D-R2N2), \cite{groueix2018papier} (AtlasNet),
from \cite{peng2021shape,boulch2022poco}. Figures \ref{fig:sn3k5} and \ref{fig:sn3k25} show visual comparisons of reconstructions from 3000 points with noises of 0.005 and 0.025 respectively. 

Our method outperforms the state-of-the-art across all metrics, and the gap increases overall with input sparsity and noise variance. Using denoising allows us in particular to improve our performance \wrt to our baseline especially in the most extreme case (sparse inputs and big variance noise). We note that our denoising-free baseline already achieves mostly on par results with the state-of-the-art method POCO, while using only half the number of parameters (Ours 6.5M, POCO 12.5M). Poisson reconstruction (SPR) is outperformed by learning based methods. Qualitative results show that our method recovers fine structures and details with more fidelity (Figure \ref{fig:sn3k5}), and that it can perform robustly even under heavy noise (Figure \ref{fig:sn3k25}), thanks to the combination of local and global reasoning at the decoder, and the denoising of the decoder's support. 

Although they offer mostly superior performance, we note that intrinsic models such as POCO and ours are less parameter efficient then extrinsic ones (\eg ConvONet and SAP contain about 2M parameters). While SAP offers a good model size/performance combination, it is however limited to small scenes by design. We note additionally that our reconstruction time at resolution $128^3$ is 538ms (POCO: 1300ms, SAP: 64ms, ConvONet: 381ms).  

\begin{table}[t!]
\centering
\scalebox{0.8}{
\begin{tabular}{lccc}
\toprule
\multicolumn{1}{c}{}  & $\text{CD}_\text{1}$ $\downarrow$ & NC $\uparrow$ & FS $\uparrow$\\
\midrule
SA-ConvONet \cite{tang2021sa}                        & 0.56   & 0.93 & 0.92\\
Neural-Pull \cite{ma2020neural}                      & 0.71 &   0.85 & 0.83\\
Ours             & \textbf{0.30} & \textbf{0.96} & \textbf{0.99}\\
\bottomrule
\end{tabular}}
\caption{Comparison to test-time fitting methods on class Tables of ShapeNet. Reconstruction from 3000 noisy (0.005 standard deviation) points.}
\label{tab:opt}
\end{table}

In Table \ref{tab:opt}, we compare our feed-forward generalizable method to deep learning optimization based approaches using their official implementations, specifically fitting an MLP to the input point cloud as in Neural-Pull \cite{ma2020neural}, and finetuning ConvONet: SA-ConvONet \cite{tang2021sa}. As optimization methods are time consuming, we follow here the slipt of Class Tables in \cite{williams2021neural}. Our method outperforms these competing methods.


\begin{figure}[t!]
  \centering
\includegraphics[width=0.7\linewidth]{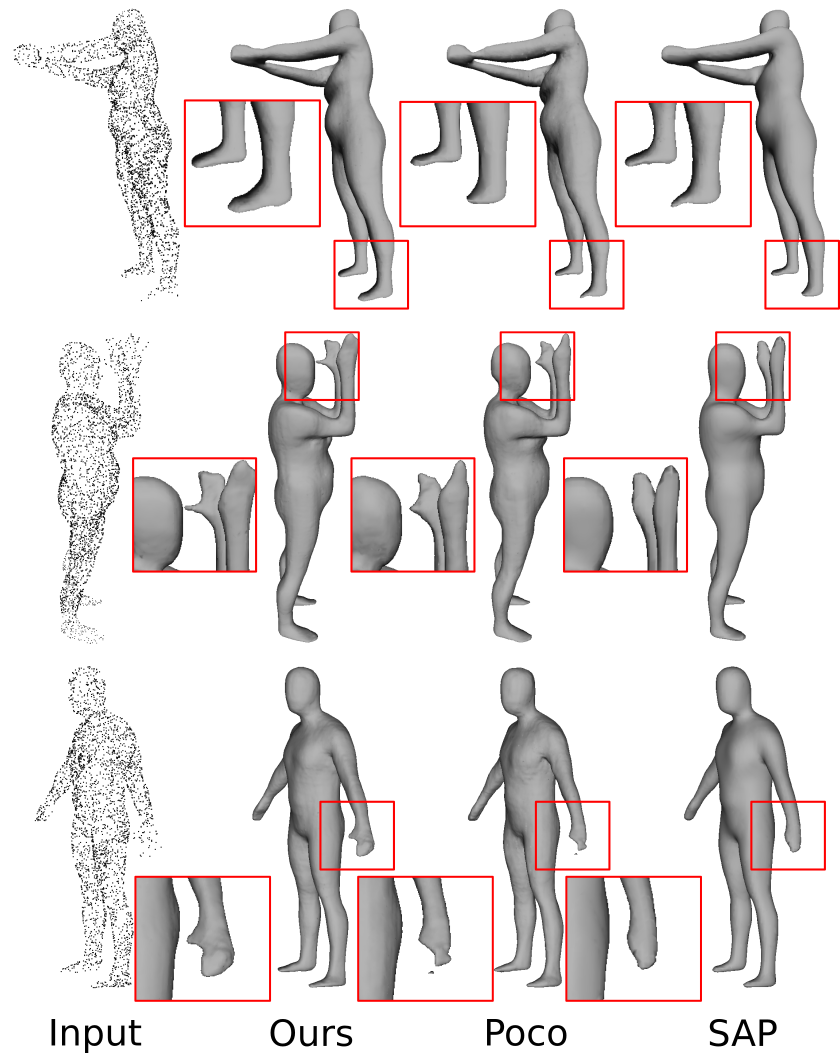}
\caption{FAUST reconstructions from real 3000 scan points. Models are \textbf{trained on ShapeNet}, and they were \textbf{not trained on any human or articulated shapes}.}
\label{fig:fst}
\end{figure}


\begin{table}[t!]
\centering
\scalebox{0.8}{
\begin{tabular}{lcccc}
\toprule
\multicolumn{1}{c}{} & $\text{CD}_\text{1}$ $\downarrow$ & $\text{CD}_\text{2}$ $\downarrow$ & NC $\uparrow$ & FS $\uparrow$\\
\midrule
SAP \cite{peng2021shape}                & 0.29 & 0.13  & 0.95 & 0.98\\
POCO \cite{boulch2022poco}           & 0.25 & 0.09 & \textbf{0.96} & \textbf{0.99}\\
Ours            & \textbf{0.23} & \textbf{0.08} & \textbf{0.96} & \textbf{0.99}\\
\bottomrule
\end{tabular}}
\caption{FAUST reconstruction from 3000 points.}
\label{tab:fr}
\end{table}

\subsection{Object to Real Articulated Shape Generalization}
To assess the capacity of our model to generalize outside the training shape distribution (ShapeNet), as well as to reconstruct from real scans, we task the ShapeNet trained model with reconstructing FAUST models from 3000 points. Table \ref{tab:fr} shows numerical comparisons using the real scans, with qualitative comparisons displayed in Figure \ref{fig:fst}. We report numbers for POCO and SAP using their available ShapeNet models. While all presented methods generalize relatively successfully despite not being trained on any articulated shapes, we outperform state-of-the-art methods POCO and SAP on most metrics.

\subsection{Object to Real Scenes Generalization}
\begin{figure}[t!]
  \centering
\includegraphics[width=1.0\linewidth]{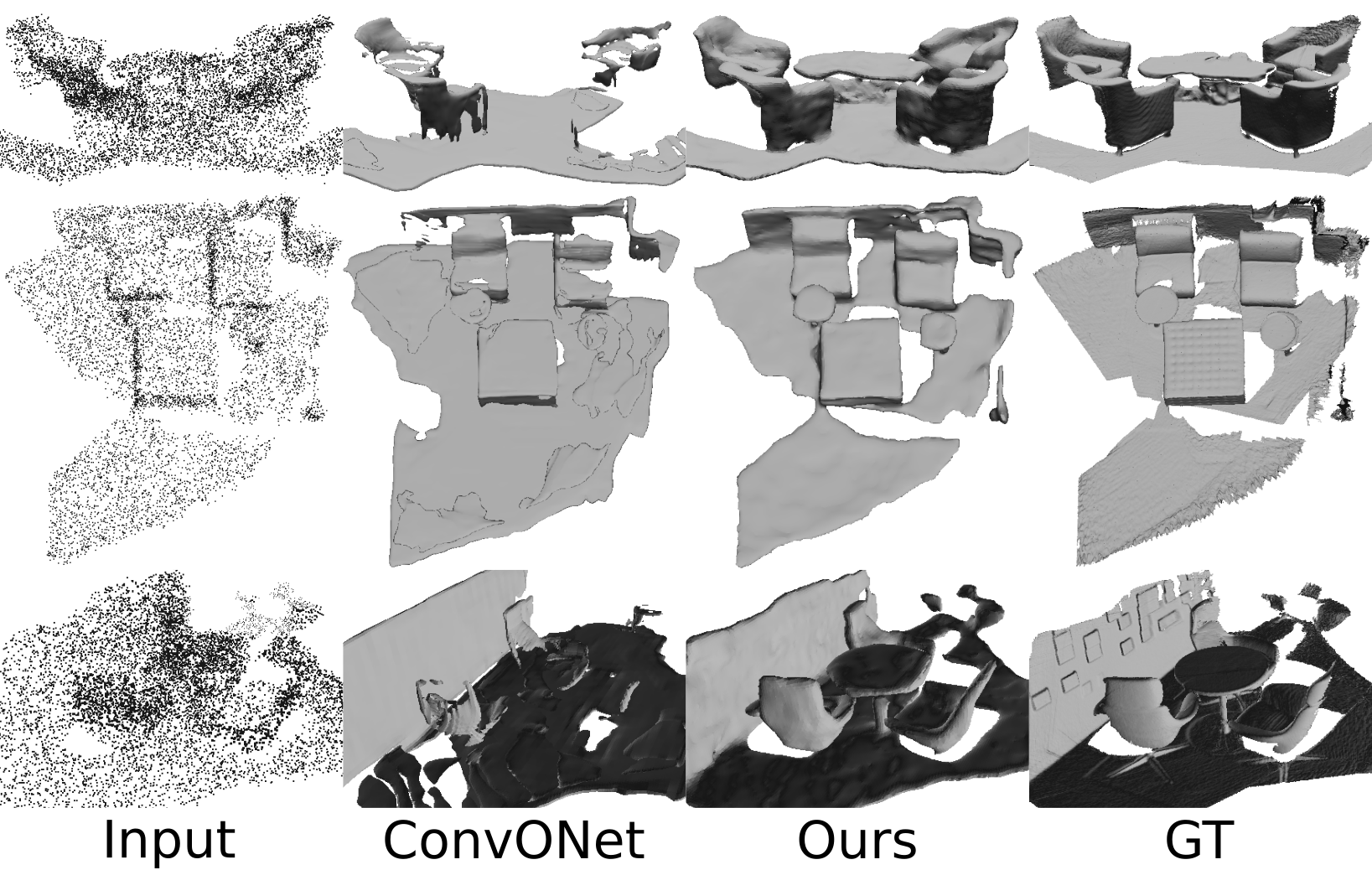}
\caption{ScanNet reconstructions from real 10k depth sensor points.}
\label{fig:scan}
\end{figure}

We extend generalization experiments to a more challenging scenario. We use both our's, SAP's and POCO's ShapeNet models trained with 3k sized inputs to reconstruct the real scans of ScanNet \cite{dai2017scannet} using 10k sized inputs. We report results for ConvONet from their paper as a reference point. Table \ref{tab:scan} shows numerical results where we outperform the competition. Figure \ref{fig:scan} shows a qualitative comparison. We note that as the ConvONet model we show was trained on the Synthetic Rooms dataset using 10k inputs, it tends to produce more flat and extrapolated planar surfaces. Although trained only on objects, our model displays satisfactory scene generalization. We notice also that it tends to be more faithful to the input. These results raise the question of extrapolation vs. input fidelity trade-off desired from reconstruction models, which we would like to explore in future work.

\begin{table}[t!]
\centering
\scalebox{0.8}{
\begin{tabular}{lcc}
\toprule
\multicolumn{1}{c}{} & $\text{CD}_\text{1}$ $\downarrow$ & FS $\uparrow$\\
\midrule
ConvONet \cite{peng2020convolutional} & 0.77 & 0.89 \\
\midrule
SAP \cite{peng2021shape} & 1.11  & 0.71 \\
POCO \cite{boulch2022poco} & 1.03  & 0.72 \\
\midrule
Ours w/o denois. & 0.69  & 0.86 \\
Ours & \textbf{0.58} & \textbf{0.91}  \\ 
\bottomrule
\end{tabular}}
\caption{ScanNet reconstructions from 10k points.}
\label{tab:scan}
\end{table}

\subsection{Ablation}
\label{sec:abl}

We propose a quantitative analysis of the impact of the components of our method at this junction on class Tables of ShapeNet. Results are summarized in Table \ref{tab:abl}. Starting from our full method (Ours), we ablate denoising (Ours w/o den.) then the global decoding (Ours w/o den. w/o glob.). We also show numbers for POCO. The global decoder helps robustify the method especially against outliers as witnessed by the improvement in L2 chamfer distance when the global mixing is introduced. Denoising improves performance consistently across all metrics as also confirmed in the previous experiments. 

\begin{table}[t!]
\centering
\scalebox{0.8}{
\begin{tabular}{lccccc}
\toprule
\multicolumn{1}{c}{} & $\text{CD}_\text{1}$ $\downarrow$ & $\text{CD}_\text{2}$ $\downarrow$ & NC $\uparrow$ & FS $\uparrow$\\
\midrule
 POCO \cite{boulch2022poco}                          & 0.36 & 0.23 & 0.95 & 0.97 \\
\midrule
 Ours\small{ w/o den. w/o glob.} & 0.34 & 0.23 & 0.96 & 0.98 \\
 Ours\small{ w/o den.}          & 0.33 & 0.19 & 0.96 & 0.98 \\
 Ours                            & \textbf{0.31} & \textbf{0.14} & \textbf{0.97} & \textbf{0.99} \\
\bottomrule
\end{tabular}}
\caption{Ablation on class Tables of ShapeNet. Reconstruction from 3000 noisy (0.005 std. dev.) points.}
\label{tab:abl}
\end{table}

\section{Conclusion}

We showed in this work that coupling denoising with reconstruction is beneficial for intrinsic implicit feed forward reconstruction models. We also showed that for these types of models, a fully MLP based architecture can produce state-of-the-art results with less parameters compared to a convolutional counterpart. This result questions the utility of convolution based locality mechanisms in this context.

\bibliographystyle{ieeenat_fullname}
\bibliography{main}

\input{supp_sec}


\end{document}

%% file: preamble.tex
\usepackage[dvipsnames]{xcolor}


%% file: supp_sec.tex
\clearpage
\setcounter{page}{1}
\maketitlesupplementary

\section{Evaluation metrics}
Following the definitions from \cite{boulch2022poco}, we present, in this section, the formal definitions for the metrics that we use for evaluation. We denote by $\mathcal{S}$ and  $\hat{\mathcal{S}}$ the ground truth and predicted mesh respectively. All metrics are approximated with 100k samples from $\mathcal{S}$ and  $\hat{\mathcal{S}}$. 

$$
$$

\paragraph{Chamfer distance}
We provide two variations of the Chamfer distance using the two-ways nearest neighbor distance and squared distance, respectively: 
$$\mathrm{CD}_1=\frac{1}{2|\mathcal{S}|} \sum_{v \in \mathcal{S}} \min _{\hat{v} \in \hat{\mathcal{S}}}\|v-\hat{v}\|_2+\frac{1}{2|\hat{\mathcal{S}}|} \sum_{\hat{v} \in \hat{\mathcal{S}}} \min _{v \in \mathcal{S}}\|\hat{v}-v\|_2$$

$$\mathrm{CD}_2=\frac{1}{2|\mathcal{S}|} \sum_{v \in \mathcal{S}} \min _{\hat{v} \in \hat{\mathcal{S}}}\|v-\hat{v}\|_2^2+\frac{1}{2|\hat{\mathcal{S}}|} \sum_{\hat{v} \in \hat{\mathcal{S}}} \min _{v \in \mathcal{S}}\|\hat{v}-v\|_2^2$$

\paragraph{F-Score (FS)} For a given threshold $\tau$, the F-score between  the meshes $\mathcal{S}$ and $\hat{\mathcal{S}}$ is defined as:
$$
\mathrm{FS}\left(\tau, \mathcal{S}, \hat{\mathcal{S}}\right)=\frac{2 \text { Recall} \cdot \text{Precision }}{\text { Recall }+\text { Precision }}
$$

where
$$
\begin{array}{r}
\operatorname{Recall}\left(\tau, \mathcal{S}, \hat{\mathcal{S}}\right)=\mid\left\{v \in \mathcal{S} \text {, s.t. } \min _{\hat{v} \in \hat{ \mathcal{S} }} \left\|v-\hat{v}\|_2\right<\tau\right\} \mid \\
\operatorname{Precision}\left(\tau, \mathcal{S}, \hat{\mathcal{S}}\right)=\mid\left\{\hat{v} \in \hat{\mathcal{S} }\text {, s.t. } \min _{v \in  \mathcal{S} } \left\|v-\hat{v}\|_2\right<\tau\right\} \mid \\
\end{array}
$$
Following ONet \cite{mescheder2019occupancy} and ConvONet \cite{peng2020convolutional}, we set $\tau$ to  $0.01$.

\paragraph{Normal consistency (NC)} We denote here by $n_v$ the normal at a point $v$ in $\mathcal{S}$. The normal consistency between two meshes $\mathcal{S}$ and $\hat{\mathcal{S}}$ is defined as: 

$$\mathrm{NC}=\frac{1}{2|\mathcal{S}|} \sum_{v \in \mathcal{S}} n_{v} \cdot n_{\operatorname{closest}\left(v,\hat{ \mathcal{S}}\right)}+\frac{1}{2|\hat{\mathcal{S}}|} \sum_{\hat{v} \in \hat{\mathcal{S}}} n_{\hat{v}} \cdot n_{\operatorname{closest}\left(\hat{v}, \mathcal{S}\right)}$$

where  
$$
\operatorname{closest}(v, \hat{\mathcal{S}}) = \operatorname{argmin} _{\hat{v} \in \hat{\mathcal{S}}}\|v-\hat{v}\|_2
$$

\section{Additional Object Reconstruction Numerical Results}
We report our per class ShapeNet\cite{shapenet} reconstruction results  from 3000 noisy points in Table \ref{tab:3k}, and 300 noisy points in Table \ref{tab:300}.  We show numbers for both noises of 0.005 and 0.025 standard deviation.

\begin{table}[h!]
\centering
\begin{tabular}{ccccc}
\toprule
                                                                & \multicolumn{2}{c}{$\sigma$ = 0.005}                                   & \multicolumn{2}{c}{$\sigma$ = 0.025}                                                                                                                                 \\ \midrule 
                                                                  & \multicolumn{1}{c}{$\text{CD}_\text{1}$ $\downarrow$}  & $\text{CD}_\text{2}$ $\downarrow$                            & \multicolumn{1}{c}{$\text{CD}_\text{1}$ $\downarrow$}                          & \multicolumn{1}{c}{$\text{CD}_\text{2}$ $\downarrow$}                          \\ \midrule %
\multicolumn{1}{l}{Airplane}  & \multicolumn{1}{c}{0.23} & 0.08 & \multicolumn{1}{c}{0.48} & \multicolumn{1}{c}{0.42} \\ 
\multicolumn{1}{l}{Bench}                             & \multicolumn{1}{c}{0.29} & 0.12                          & \multicolumn{1}{c}{0.52}                         & \multicolumn{1}{c}{0.51}                         \\ 
\multicolumn{1}{l}{Cabinet}   & \multicolumn{1}{c}{0.35} & 0.19 &  \multicolumn{1}{c}{0.56} & \multicolumn{1}{c}{0.91} \\ 
\multicolumn{1}{l}{Car}                              & \multicolumn{1}{c}{0.39} & 0.31 &  \multicolumn{1}{c}{0.78}                         & \multicolumn{1}{c}{1.62}                         \\ 
\multicolumn{1}{l}{Chair}     & \multicolumn{1}{c}{0.34} & 0.17  & \multicolumn{1}{c}{0.60} & \multicolumn{1}{c}{0.73} \\ 
\multicolumn{1}{l}{Display}                           & \multicolumn{1}{c}{0.29} & 0.12 &  \multicolumn{1}{c}{0.49}                         & \multicolumn{1}{c}{0.43}                         \\ 
\multicolumn{1}{l}{Lamp}      & \multicolumn{1}{c}{0.29} & 0.19 &  \multicolumn{1}{c}{0.71} & \multicolumn{1}{c}{3.90} \\ 
\multicolumn{1}{l}{Speaker}                           & \multicolumn{1}{c}{0.40} & 0.32 &  \multicolumn{1}{c}{0.67}                         & \multicolumn{1}{c}{1.18}                         \\ 
\multicolumn{1}{l}{Rifle}     & \multicolumn{1}{c}{0.19} & 0.05 & \multicolumn{1}{c}{0.44} & \multicolumn{1}{c}{0.37} \\ 
\multicolumn{1}{l}{Sofa}                              & \multicolumn{1}{c}{0.30} & 0.14 &  \multicolumn{1}{c}{0.53}                         & \multicolumn{1}{c}{0.56}                         \\ 
\multicolumn{1}{l}{Table}     & \multicolumn{1}{c}{0.31} & 0.14 &  \multicolumn{1}{c}{0.50} & \multicolumn{1}{c}{0.59} \\ 
\multicolumn{1}{l}{Phone}                             & \multicolumn{1}{c}{0.23} & 0.07 &  \multicolumn{1}{c}{0.36}                         & \multicolumn{1}{c}{0.22}                         \\ 
\multicolumn{1}{l}{Vessel}    & \multicolumn{1}{c}{0.25} & 0.10 &  \multicolumn{1}{c}{0.57} & \multicolumn{1}{c}{0.68} \\ 
\bottomrule
\end{tabular}
\caption{ShapeNet reconstruction from 3000 noisy points, with standard deviations ($\sigma$) of 0.005 and 0.025.}
\label{tab:3k}
\end{table}

\begin{table}[h!]
\centering

\begin{tabular}{ccccc}
\toprule
                                                                & \multicolumn{2}{c}{$\sigma$ = 0.005}                                   & \multicolumn{2}{c}{$\sigma$ = 0.025}                                                                                                                                 \\ \midrule
                                                                & \multicolumn{1}{c}{$\text{CD}_\text{1}$ $\downarrow$}  & $\text{CD}_\text{2}$ $\downarrow$  &  \multicolumn{1}{c}{$\text{CD}_\text{1}$ $\downarrow$}                          & \multicolumn{1}{c}{$\text{CD}_\text{2}$ $\downarrow$}                          \\ \midrule
\multicolumn{1}{l}{Airplane} &  \multicolumn{1}{c}{0.45} & 0.45  & \multicolumn{1}{c}{0.89} & 1.88                                             \\ 
\multicolumn{1}{l}{Bench}    & \multicolumn{1}{c}{0.50} & 0.56  & \multicolumn{1}{c}{0.97} & 2.70                                             \\ 
\multicolumn{1}{l}{Cabinet}  & \multicolumn{1}{c}{0.56} & 0.88 & \multicolumn{1}{c}{0.95} & 2.26                                             \\ 
\multicolumn{1}{l}{Car}      & \multicolumn{1}{c}{0.83} & 2.02 & \multicolumn{1}{c}{1.21} & 3.20                                             \\ 
\multicolumn{1}{l}{Chair}    & \multicolumn{1}{c}{0.62} & 0.97 & \multicolumn{1}{c}{1.21} & 5.50                                             \\ 
\multicolumn{1}{l}{Display}  & \multicolumn{1}{c}{0.48} & 0.54 & \multicolumn{1}{c}{0.89} & 1.63                                             \\ 
\multicolumn{1}{l}{Lamp}     & \multicolumn{1}{c}{0.67} & 2.32  & \multicolumn{1}{c}{1.39} & 14.17                                            \\ 
\multicolumn{1}{l}{Speaker}  & \multicolumn{1}{c}{0.75} & 2.03  & \multicolumn{1}{c}{1.26} & 4.34                                             \\ 
\multicolumn{1}{l}{Rifle}    & \multicolumn{1}{c}{0.36} & 0.26  & \multicolumn{1}{c}{0.72} & 1.17                                             \\ 
\multicolumn{1}{l}{Sofa}     & \multicolumn{1}{c}{0.52} & 0.64 & \multicolumn{1}{c}{1.00} & 2.09                                             \\ 
\multicolumn{1}{l}{Table}    & \multicolumn{1}{c}{0.52} & 0.75  & \multicolumn{1}{c}{0.95} & 3.21                                             \\ 
\multicolumn{1}{l}{Phone}    & \multicolumn{1}{c}{0.34} & 0.21 & \multicolumn{1}{c}{0.63} & 0.78                                             \\ 
\multicolumn{1}{l}{Vessel}   & \multicolumn{1}{c}{0.51} & 0.61 & \multicolumn{1}{c}{1.03} & 2.40                                             \\ \bottomrule
\end{tabular}
\caption{ShapeNet reconstruction from 300 noisy points, with standard deviations ($\sigma$) of 0.005 and 0.025.}
\label{tab:300}
\end{table}

\section{Additional Object Reconstruction Qualitative Results}
\begin{figure}[t!]
  \centering
\includegraphics[width=0.7\linewidth]{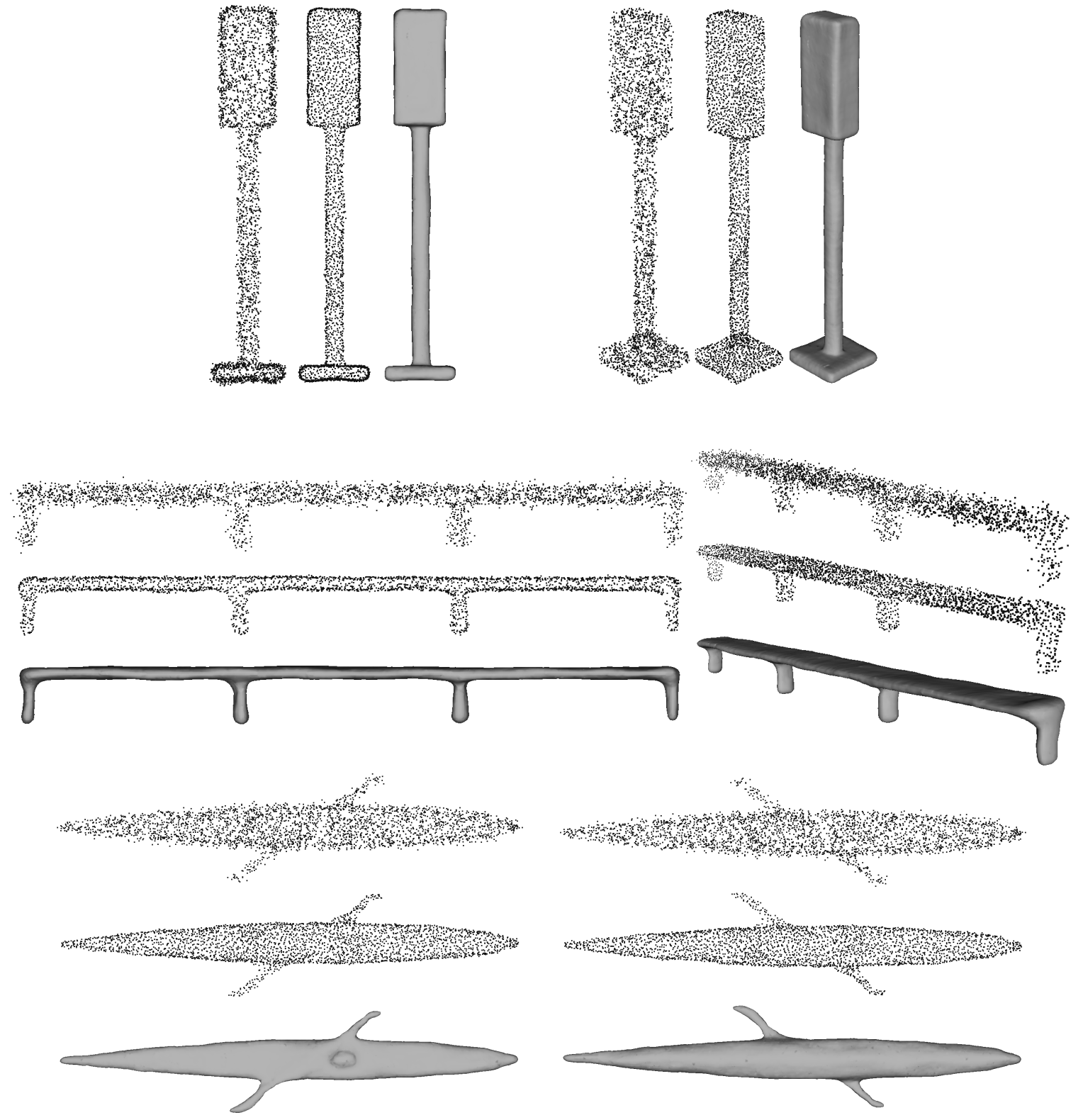}
\caption{ShapeNet reconstructions from 3000 noisy (0.005) points.
We show the \textbf{input} point cloud, our \textbf{denoising} and our \textbf{reconstruction}.}
\label{fig:q5}
\end{figure}
We show additional ShapeNet\cite{shapenet} reconstructions using our method from 3000 noisy points with noise standard deviation  0.005 in Figure \ref{fig:q5}. We show the input point cloud, our denoised point cloud, and the output mesh.

\section{Additional Scene Reconstruction Results}

\begin{figure}[h!]
  \centering
\includegraphics[width=0.7\linewidth]{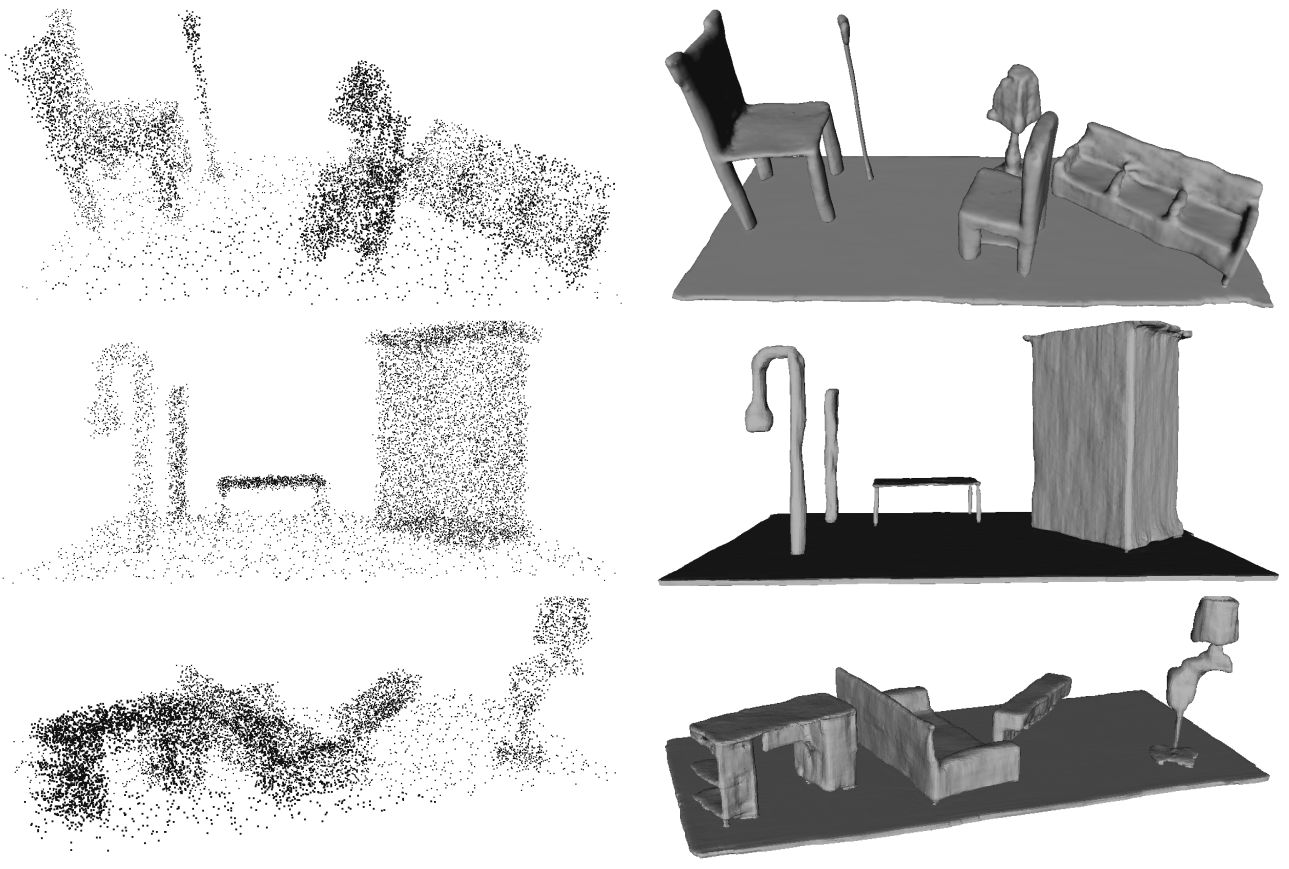}
\caption{Synthetic Rooms reconstructions from 10k noisy points, with standard deviation of 0.005.}
\label{fig:sr}
\end{figure}

\begin{table}[h!]
\centering
\scalebox{1.0}{
\begin{tabular}{lccccc}
\toprule
\multicolumn{1}{c}{} & $\text{CD}_\text{1}$ $\downarrow$ & $\text{CD}_\text{2}$ $\downarrow$ & NC $\uparrow$ & FS $\uparrow$\\
\midrule
SPR \cite{kazhdan2013screened} & 2.23 & - & 0.87 & 0.81\\ 
ONet \cite{mescheder2019occupancy} & 2.03 & - & 0.78 & 0.54\\
DP-ConvONet \cite{lionar2021dynamic} & 0.42 & - & 0.91 & 0.96 \\ 
ConvONet \cite{peng2020convolutional}            & 0.42 & - & 0.91 & 0.96 \\
POCO \cite{boulch2022poco}                 & \textbf{0.36} & 0.31 & \textbf{0.92} & \textbf{0.98} \\
\midrule
Ours\small{ w/o den.}& 0.38 & 0.29  & 0.90 & 0.97 \\
Ours                 & \textbf{0.36} & \textbf{0.25} & 0.91 & \textbf{0.98} \\
\bottomrule
\end{tabular}}
\caption{Synthetic Rooms reconstruction from 10k noisy points, with standard deviation of 0.005. \textbf{POCO: 12.5M params./ Ours: 6.5M params.}}
\label{tab:sr}
\end{table}

Table \ref{tab:sr} shows  numerical evaluations of Synthetic Rooms \cite{peng2020convolutional} reconstructions from 10k points with 0.005 standard deviation noise. We report numbers for POCO using their available model and we compile numbers for ConvONet \cite{peng2020convolutional}, DP-ConvONet \cite{lionar2021dynamic}, SPR \cite{kazhdan2013screened}, ONet \cite{mescheder2019occupancy} from \cite{boulch2022poco}. Figure \ref{fig:sr} shows a few reconstruction examples. We perform overall on par with POCO on this task, while achieving a lower L2 Chamfer score, which we believe is on account of our global decoder and denoising.  We pinpoint again that we are working with only half the model size of POCO (6.5M \vs 12.5M). 